# Enhancing Modern Supervised Word Sense Disambiguation Models by Semantic Lexical Resources

**Stefano Melacci[1,2], Achille Globo[2], Leonardo Rigutini[2]**
[1]DIISM - University of Siena, Via Roma 56 – 53100 Siena (Italy)
[2]QuestIT s.r.l., Via Leonida Cialfi 23 – 53100 Siena (Italy)
mela@diism.unisi.it, {globo, rigutini}@quest-it.com

## Abstract

Supervised models for Word Sense Disambiguation (WSD) currently yield to state-of-the-art results in the most popular benchmarks. Despite the recent introduction of Word Embeddings and Recurrent Neural Networks to design powerful context-related features, the interest in improving WSD models using Semantic Lexical Resources (SLRs) is mostly restricted to knowledge-based approaches. In this paper, we enhance "modern" supervised WSD models exploiting two popular SLRs: WordNet and WordNet Domains. We propose an effective way to introduce semantic features into the classifiers, and we consider using the SLR structure to augment the training data. We study the effect of different types of semantic features, investigating their interaction with local contexts encoded by means of mixtures of Word Embeddings or Recurrent Neural Networks, and we extend the proposed model into a novel multi-layer architecture for WSD. A detailed experimental comparison in the recent Unified Evaluation Framework (Raganato et al., 2017) shows that the proposed approach leads to supervised models that compare favourably with the state-of-the art.

## 1. Introduction

Determining the correct meaning of a target word in a given context is a problem that is commonly referred to as Word Sense Disambiguation (WSD). WSD has a long tradition in the Natural Language Processing (NLP) community (Navigli, 2009), and it is still a very challenging task, subject to recent studies (Camacho-Collados et al., 2016; Raganato et al., 2017). The universe of WSD approaches is usually divided into the two main categories of "supervised" and "knowledge-based" methods (Raganato et al., 2017). The first category includes those algorithms that exploit manually annotated corpora (Zhong and Ng, 2010), and, recently, we also observe the growth of techniques that benefit from additional (semi) automatically annotated data (Taghipour and Ng, 2015a; Bovi et al., 2017) or that cast the learning problem into the semi-supervised setting (Taghipour and Ng, 2015b; Yuan et al., 2016). The second category considers those approaches that disambiguate words only using structured sources of lexical knowledge (Lesk, 1986; Moro et al., 2014; Agirre et al., 2014; Weissenborn et al., 2015). On one hand, knowledge-based systems are easier to setup, not requiring annotated corpora; on the other hand, supervised models currently yield to state-of-the-art results in the most popular benchmarks (Raganato et al., 2017).

Recent supervised WSD models are frequently based on the "It Makes Sense" (IMS) framework (Zhong and Ng, 2010), whose original implementation is also studied when augmented with Word Embeddings (WEs) to generate a distributed representations of the local context around the target word (Taghipour and Ng, 2015c; Rothe and Schütze, 2015). Other WSD approaches are based on Recurrent Neural Networks (RNNs) (Yuan et al., 2016; Melamud et al., 2016), such as the Context2Vec (C2V) model (Melamud et al., 2016), where a Bidirectional RNN is used to embed local contexts in a space where a supervised distance-based classifier performs the disambiguation.

WordNet (Princeton University, 2006) is commonly assumed to be the sense-repository for WSD. Each *word type* (lemma + Part of Speech (PoS)) is paired with one or more *senses*, and senses that express the same concept (among different word types) are grouped into a *synset*. Synsets participate to semantic/lexical relations, they contain definitions (*glosses*) and, in several cases, one or more *example sentences* (Miller, 1995). WordNet Domains (Fondazione Bruno Kessler, 2007) extends WordNet with ($\approx$ 200) semantic domain labels (Bentivogli et al., 2004).

Several WSD approaches exploit WordNet and WordNet Domains to implement *semantic features* (Magnini et al., 2001; Bakx et al., 2006; Bell and Patrick, 2004; Martinez, 2005; Kolte and Bhirud, 2008; Khapra et al., 2010; Navigli, 2012). The outcome of the feature extraction stage is often a representation of the local context of the target word based on domain vectors (Magnini et al., 2001; Bell and Patrick, 2004), collection of synsets (Martinez, 2005), and other domain-related statistics (Bakx et al., 2006; Kolte and Bhirud, 2008; Khapra et al., 2010; Navigli, 2012).

In this paper, we focus on the IMS framework and on "modern" implementations of the IMS features based on distributed representations. We propose an effective way to introduce SLR-based semantic features into the word-sense classifiers, and to possibly exploit the SLRs and their structure to augment the training data. We study the effect of different types of semantic features, and we extend the proposed models into a novel multi-layer architecture for WSD. To our best knowledge, the two aforementioned Semantic Lexical Resources (SLRs) have not been recently studied in conjunction with IMS models. A detailed experimental comparison in the Unified Evaluation Framework (Raganato et al., 2017) shows that the proposed approaches leads to supervised models that compare favourably with the state-of-the art.

## 2. Models

We are given a collection of labeled training sentences, in which, for a subset of the words, we are provided annotations on the disambiguated senses associated to them. Common supervised approaches train a classifier for each of the word types for which more than one of different sense annotations are found in the training sentences. When testing the classifiers on word types for which no-classifiers are available, a fall-back policy is employed, based on the WordNet First Sense (WFS). As a matter of fact, WordNet senses are ordered with respect to their estimated frequency, and the WFS is the most common sense for each word type. We focus on the IMS system (Zhong and Ng, 2010), where a linear Support Vector Machine (SVM) classifier is used to classify vectorial representations of the contexts around the target word. We indicate with $f_{wt}(x)$ the classifier associated to the word type $wt$, that outputs a decision over $c_{wt}$ classes/senses. In this paper, we consider several implementations of the feature space to which $x$ belongs. In particular, we partition $x$ as follows

$$x = \left[ \underbrace{\overbrace{x_{PoS}, x_{locCol}, x_{sWords}}^{x_{IMS}}, x_{contextEmb}}_{x_{IMSWE} \text{ or } x_{IMSC2V}}, x_{sem} \right], \quad (1)$$

where the comma represent the vector concatenation operator. Vector $x_{IMS}$ corresponds to the default implementation of IMS, composed of PoS-tag features $x_{PoS}$ (in a window of size 7 centered in the target word), local collocations $x_{locCol}$ (11 local collocations), and binary indicators of the "surrounding words" in the current context $x_{sWords}$ (the context is limited to the sentence to which the target word belong and, eventually, some other sentences before and after it), refer to (Zhong and Ng, 2010) for more details. A distributed representation of the context around the target word, $x_{contextEmb}$, is implemented by means of weighted mixtures of Word Embeddings (WE) (we follow Iacobacci et al. (2016), and we use exponentially weighted sums of WEs, computed with Word2Vec (Mikolov et al., 2013)) or Bidirectional RNNs in the C2V approach (Melamud et al., 2016), leading to the input representations $x_{IMSWE}$ and $x_{IMSC2V}$, respectively. The WE model[1] and the C2V net[2] that we consider in this paper are trained in the English ukWaC corpus (Baroni et al., 2009), composed of up of two billion words (see the websites in the footnotes for all the details). Strictly speaking, including these features make the whole classifier semi-supervised, since we are actually using an additional, unlabeled resource (ukWaC). Under some circumstances, good results have been obtained by discarding $x_{sWords}$ when $x_{contextEmb}$ is present (Iacobacci et al., 2016). In our experience, this effect becomes evident when multiple training sentences are considered to build $x_{sWords}$, while here we focus on a single sentence. We notice that, to our best knowledge, this way of including the C2V net into the IMS framework was not experimented before, while it makes the comparisons more uniform and, as we will show later, it leads to improved results.

We propose to compute also a set of semantic features $x_{sem}$ to generate a more informative representation of the local context of the target word. To this extent, we make use of WordNet, extended with the information from WordNet Domains. We design three different types of semantic features, referred to as "PR", "sSyn", and "*Dom*", and we consider also every combination of them to generate and evaluate multiple instances of $x_{sem}$. These features are computed on disambiguated words, so that prior disambiguation hypotheses are needed (we will shortly return on this point). In detail,

- "PR" models prior information, and it is a 1-hot encoding of the most likely sense for the target word in the current context.

- "sSyn" is the collection of "surrounding synsets" that appear in the context of the target word (i.e., in the same sentence). They are represented as binary indicators analogously to the surrounding words of IMS (Zhong and Ng, 2010), but, differently to them, they are a less variable representation of the context, since synonyms are represented with a single symbol.

- "*Dom*" encodes the 3 most frequent domains of the WordNet Domains taxonomy that are found in the context of the target word (they are encoded as binary features and in the case of "sSyn"), discarding the "factotum" domain. Domain information is known to help the WSD process (Magnini et al., 2001).

In order compute these features, we would need to have the use of other WSD models, while here we exploit the idea of using the WFS heuristic to disambiguate the context and compute priors on the target word. WFS has been show to be a competitive and hard-to-beat baseline (Raganato et al., 2017), even if its computational cost is almost negligible. As we will see shortly, the "PR" feature gains more importance in the multi-layer setting. In single-layer models, this feature is constant among all the training instances, and the SVM classifier will end-up in developing a bias term regularized by a squared penalty (differently from the usual unregularized SVM bias).

One of the crucial issues of supervised WSD is the lack of large collections of training examples. In this work we explore two ways of augmenting the training sets by means of WordNet. Word types belonging to same synset are synonyms, and this structure can be used to augment the training data in an efficient way: in order to train $f_{wt}$, we inherit all the training examples associated to other word types/senses sharing synsets with the target senses of $wt$. We talk about classifiers augmented using Synset Level Information (SIL). Another useful resource of training examples are WordNet glosses and example sentences. It is pretty trivial to automatically disambiguate most of the example sentences, since the word types that they describe are known. We notice that many knowledge-based systems make use of another linguistic resource, that is the Princeton WordNet Gloss Corpus, a collection of manually (and automatically) disambiguated definitions of the gloss data[3]. We investigate the use of this resource in supervised WSD.

---

[1] http://lcl.uniroma1.it/wsdeval/systems
[2] http://u.cs.biu.ac.il/~nlp/resources/downloads/context2vec/
[3] http://wordnet.princeton.edu/glosstag.shtml

Another direction we explore is when the semantic features are computed within a multi-layer architecture. In other words, we use the trained WSD model described so far to disambiguate the context of the target words and compute priors on their senses (a related idea was exploited, for example, in (Khapra et al., 2010)), then we update the semantic features and train another model. This procedure can be repeated multiple times. Formally, if we indicate with $\ell \geq 1$ the layer index, we have

$$f_{wt}^{\ell}\left(\boldsymbol{x}^{\ell}\right) = f_{wt}^{\ell}\left(\left[\boldsymbol{x}_{IMS*},\ \boldsymbol{x}_{sem}^{\ell-1}\right]\right)\ ,$$

where IMS* is one of {IMS, IMSWE, IMSC2V}, $\boldsymbol{x}_{sem}^{\ell-1}$ is computed using the disambiguation hypotheses given by $f_{wt}^{\ell-1}$, and $\boldsymbol{x}_{sem}^{0}$ is based on the WFS. This architecture must be trained in a layer-wise fashion (train layer $\ell$, disambiguate training data, train layer $\ell+1$, etc.), but we also explore the simpler case in which we only train $f_{wt}^{1}$ and, at test time, we iteratively apply it in a multi-layer scheme.

### 3. Experimental Results

We compare three main models, IMS, IMSWE, IMSC2V, respectively (based on the related input representations of Eq. 1), with and without semantic features, in the recently proposed Unified Evaluation Framework (Raganato et al., 2017), that includes 5 popular benchmarks (Senseval2, Senseval3, SemEval2007, SemEval2013, SemEval2015), and the concatenation of the sentences of all of them (ALL). Such framework provides code for computing the F1 score of the models, together with carefully processed training and test data, using WordNet 3.0 as sense repository. The training data consists in the SemCor corpus, the sense-tagged corpus created by the WordNet Project research team. We consider 5 categories of experiments, focussing on different topics (in bold).

**Semantic features.** The proposed features have different effects in the three studied systems, as reported in Table 1. They almost always provide improvements over the baseline approaches, even if in different configurations. In particular, in the IMS case, semantic features lead to more evident benefits w.r.t. what happens in the IMSWE and IMSC2V cases, since WEs and C2V already capture several word regularities. We also notice that the use of domain-related information seems to carry the most useful information for the WSD task. On average, the joint use of all the semantic features (+sSyn+PR+*Dom*) provides good improvements over all the three systems. Comparing our results with the best results (considering several supervised and knowledge-based systems) collected in (Raganato et al., 2017), our approach leads to state-of-the-art results both in the cases of IMSWE and IMSC2V. It is interesting to evaluate that the C2V features applied in the IMS framework lead to better results than in the original distance-based approach proposed in (Melamud et al., 2016) and evaluated also in (Raganato et al., 2017). Table 2 provides a detail of the experimental results considering different word classes (ALL benchmark). Semantic features allows the models to gain improvements when disambiguating adjectives and, more importantly, when disambiguating verbs, that are the most polysemous elements. This more evident in the case of IMSWE, confirming that semantic features introduce useful information that is not captured by WEs.

| | Senseval2 | Senseval3 | SemEval2007 | SemEval2013 | SemEval2015 | ALL |
|---|---|---|---|---|---|---|
| IMS | 70.2 | 68.8 | **62.2** | 65.3 | 69.3 | 68.1 |
| +PR | 70.4 | 68.8 | **62.2** | 65.1 | 69.4 | 68.2 |
| +sSyn | 70.2 | 69.4 | 61.8 | 65.0 | 69.2 | 68.1 |
| +sSyn+PR | 70.1 | 69.3 | 61.8 | 65.0 | 69.5 | 68.1 |
| +*Dom* | 70.7 | **69.8** | 60.7 | **65.4** | 69.6 | **68.5** |
| +PR+*Dom* | 70.6 | 69.7 | 60.7 | 65.0 | 69.6 | 68.3 |
| +sSyn+*Dom* | **70.8** | 69.7 | 60.9 | **65.4** | 69.5 | **68.5** |
| +sSyn+PR+*Dom* | **70.8** | 69.7 | 60.9 | 65.0 | **69.8** | 68.4 |
| IMSWE | 72.2 | 69.9 | 62.9 | **66.2** | 71.9 | 69.6 |
| +PR | 72.0 | 69.7 | 62.9 | 66.1 | 72.0 | 69.5 |
| +sSyn | 72.5 | 70.1 | 62.6 | 66.1 | 71.9 | 69.7 |
| +sSyn+PR | 72.5 | 70.1 | 62.9 | 66.1 | **72.2** | 69.8 |
| +*Dom* | **72.7** | **70.3** | 62.9 | 66.1 | 72.0 | **69.9** |
| +PR+*Dom* | 72.5 | 70.0 | **63.3** | 66.0 | 72.1 | 69.8 |
| +sSyn+*Dom* | 72.6 | 70.2 | **63.3** | 66.0 | 71.8 | 69.8 |
| +sSyn+PR+*Dom* | **72.7** | 70.1 | **63.3** | 66.1 | 71.9 | 69.8 |
| IMSC2V | 73.8 | **71.9** | 63.3 | 68.1 | 72.7 | 71.1 |
| +PR | 73.8 | **71.9** | 63.3 | **68.2** | **72.8** | **71.3** |
| +sSyn | **74.2** | 71.8 | 63.5 | 68.1 | **72.8** | **71.3** |
| +sSyn+PR | 74.1 | 71.6 | 63.3 | 68.1 | **72.8** | **71.3** |
| +*Dom* | 73.9 | 71.8 | 63.7 | 68.0 | 72.3 | 71.2 |
| +PR+*Dom* | 73.9 | **71.9** | 63.5 | 67.9 | 72.7 | 71.2 |
| +sSyn+*Dom* | 74.0 | 71.8 | **64.0** | 67.9 | 72.5 | 71.2 |
| +sSyn+PR+*Dom* | 74.0 | 71.8 | 63.7 | 67.9 | 72.6 | 71.2 |
| (Raganato et al., 2017) | 72.2 | 70.4 | 62.6 | 67.3 | 71.9 | 69.6 |

Table 1: F1 score of IMS, IMSWE, IMSC2V, and their semantic-feature-based variants. The last row collects the best results (over several models) in (Raganato et al., 2017).

**Bias toward WFS.** We investigated the tendency of the WSD system to be biased toward the most frequent sense in the training data (Postma et al., 2016), that, in our experimental setting, is very similar to the WFS. Table 3 indicates the percentage of correctly disambiguated instances for which the correct sense is *not* the WFS. The result shows that adding the PR feature does not produce significant changes. When considering all the semantic features we observe slight variations, mostly concentrated in the IMSC2V case. Since IMSC2V is the less-biased model, this stronger effect was expected. Interestingly, in the case of IMS, the best combination of semantic features leads to the best results and to a less-biased classifier.

**Augmenting the training data.** A key point to evaluate is whether the semantic features can benefit by the augmentation procedures described in Section 2 (SLI, glosses, example sentences). Table 4 shows that using the synset-based procedure (SLI), paired with semantic features, improves the classifiers, while discarding such features results in reduced scores. This is explained by the fact that the semantic features are more robust and corse-grained than the other standard IMS features, and they can handle the larger variability (and some noise) in the training data that is introduced by the augmentation procedure. A more evident improvement can be observed when introducing WordNet glosses and examples into the training set (Fig 1). Models equipped with semantic features constantly benefit from such data. Unsupervised example sentences (Fig 1(a)) cause improvements that overcome the results of Table 1. When disambiguated glosses are used (Fig 1(b-c)), the improvements are more evident, since we are actually exploiting new manually annotated data. We can also train classifiers for new word types, or expand the number of senses covered by the already existing classifiers (Fig 1(c)).

|        | Noun | Adj  | Verb | Adv  |
|--------|------|------|------|------|
| IMS    | 70.0 | 75.2 | 56.0 | **83.2** |
| +PR    | 70.0 | 75.3 | 56.0 | **83.2** |
| +sSyn  | 69.9 | 75.5 | **56.3** | 82.7 |
| +sSyn+PR | 70.0 | 75.4 | 56.1 | 82.7 |
| +*Dom* | **70.4** | **76.3** | 55.9 | 82.9 |
| +PR+*Dom* | 70.3 | **76.3** | 55.6 | 82.9 |
| +sSyn+*Dom* | **70.4** | 76.1 | 56.1 | 82.9 |
| +sSyn+PR+*Dom* | 70.3 | **76.3** | 55.9 | 83.2 |
| IMSWE  | 71.8 | 76.1 | 57.4 | **83.5** |
| +PR    | 71.7 | 75.9 | 57.3 | **83.5** |
| +sSyn  | 71.9 | 76.2 | 57.6 | 83.2 |
| +sSyn+PR | 71.9 | 76.3 | 57.7 | 83.2 |
| +*Dom* | **72.0** | 76.4 | **57.8** | 83.2 |
| +PR+*Dom* | 71.8 | 76.4 | **57.8** | 83.2 |
| +sSyn+*Dom* | 71.8 | **76.5** | **57.8** | 83.2 |
| +sSyn+PR+*Dom* | 71.8 | **76.5** | **57.8** | 83.2 |
| IMSC2V | **73.1** | 77.0 | 60.5 | 83.5 |
| +PR    | **73.1** | 77.1 | **60.6** | 83.5 |
| +sSyn  | **73.1** | 77.5 | 60.4 | **83.8** |
| +sSyn+PR | **73.1** | 77.3 | 60.2 | **83.8** |
| +*Dom* | **73.1** | 77.3 | 60.0 | 83.5 |
| +PR+*Dom* | **73.1** | 77.4 | 60.2 | 83.5 |
| +sSyn+*Dom* | 73.0 | **77.6** | 60.4 | **83.8** |
| +sSyn+PR+*Dom* | 73.0 | 77.4 | 60.5 | 83.5 |
| (Raganato et al., 2017) | 72.0 | 77.2 | 57.6 | 84.7 |

Table 2: F1 score restricted to specific PoS ("ALL" benchmark). The last row is the best results (over several different models) in (Raganato et al., 2017).

|        | Plain | +PR  | +sSyn+PR+*Dom* | Best of Tab. 1 |
|--------|-------|------|----------------|----------------|
| IMS    | 9.4   | 9.2  | 9.4            | **9.7**        |
| IMSWE  | **11.0** | **11.0** | 10.8       | 10.8           |
| IMSC2V | **13.3** | **13.3** | 12.7       | 12.8           |

Table 3: The % of correct disambiguations where the right sense is NOT the WFS ("ALL"). Last column refers to the best variants in Table 1 (+sSyn+*Dom*,+*Dom*,+sSyn, resp.).

**Multi-Layer architectures.** Fig 2 compares multi-layer models (up to 4 layers), distinguishing among models with/without augmented training sets (SLI + example sentences, i.e., no manual annotations), models that are trained in a layer-wise fashion, and those that are trained only once and tested by executing them over multiple layers. The last-mentioned models can only slightly benefit from the multi-layer execution, while the layer-wise training procedure leads to better models (w.r.t. to the single-layer case). Augmenting the training data helps such training procedure. IMSC2V is not improved by the multi-layer setting, mostly because the C2V features are very informative and single layer architectures are already powerful.

|        | Plain | | +sSyn+PR+*Dom* | |
|--------|-------|-------|----------------|--------|
|        | w/o SLI | w/ SLI | w/o SLI | w/ SLI |
| IMS    | 68.1 | 66.8 | 68.4 | **68.6** |
| IMSWE  | 69.6 | 68.7 | **69.8** | 69.7 |
| IMSC2V | 71.2 | 70.8 | 71.2 | **71.6** |

Table 4: Augmenting the training data by Synset Level Information (SLI), for plain models and models equipped with all the semantic features (F1 score, "ALL" bench).

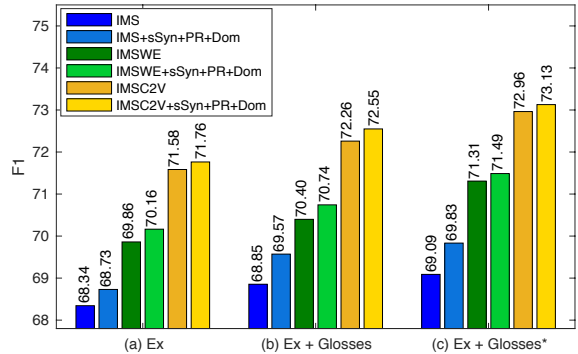

Figure 1: Training set augmentation using examples sentences only (a) and also manually disambiguated glosses (b,c), on plain models and on models equipped with all the semantic features. While (a) and (b) are limited to the word-types/senses defined in the SemCor training set, (c) considers a superset of them, using all the available annotations.

## 4. Conclusions and Future Work

We proposed to augment recent WSD models based on the IMS framework with semantic features from popular Semantic Lexical Resources: WordNet and WordNet Domains. Our deep experimental comparison shows that these features can be paired with context representations based on

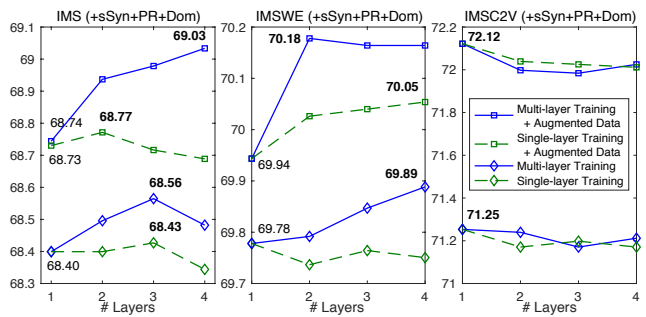

Figure 2: Multi-layer models with all the semantic features ("ALL" benchmark, F1 score): trained "layer-wise" (blue) or trained as single-layer models and iteratively ran over multiple-layers (green). Models with augmented training data exploit SLI and example sentences from WordNet.

Word Embeddings and Recurrent Neural Networks, generating more robust models. WordNet organization and data (glosses and example sentences) were used to augment the training set, showing that semantic features play a crucial role to gain enhancements. Finally, the same features are implemented into a multi-layer architecture that can improve the WSD models. Overall, we reached results that compare favourably with the state-of-the art. Future work will consider more specific semi-supervised approaches and the use of automatically annotated resources.

## 5. Bibliographical References

Agirre, E., de Lacalle, O. L., and Soroa, A. (2014). Random walks for knowledge-based word sense disambiguation. *Computational Linguistics*, 40(1):57–84.

Bakx, G. E., Villodre, L., and Claramunt, G. (2006). *Machine learning techniques for word sense disambiguation*. Ph.D. thesis, Universitat Politecnica de Catalunya.

Baroni, M., Bernardini, S., Ferraresi, A., and Zanchetta, E. (2009). The wacky wide web: a collection of very large linguistically processed web-crawled corpora. *Language resources and evaluation*, 43(3):209–226.

Bell, D. and Patrick, J. (2004). Using wordnet domains in a supervised learning word sense disambiguation system. In *Proceedings of the Australasian Language Technology Workshop 2004*, pages 17–24.

Bentivogli, L., Forner, P., Magnini, B., and Pianta, E. (2004). Revising the wordnet domains hierarchy: semantics, coverage and balancing. In *Proceedings of the Workshop on Multilingual Linguistic Resources*, pages 101–108. Association for Computational Linguistics.

Bovi, C. D., Camacho-Collados, J., Raganato, A., and Navigli, R. (2017). Eurosense: Automatic harvesting of multilingual sense annotations from parallel text. In *Proceedings of the 55th Annual Meeting of the Association for Computational Linguistics (Volume 2: Short Papers)*, volume 2, pages 594–600.

Camacho-Collados, J., Bovi, C. D., Raganato, A., and Navigli, R. (2016). A large-scale multilingual disambiguation of glosses. In *Proceedings of the Tenth International Conference on Language Resources and Evaluation (LREC 2016)*.

Iacobacci, I., Pilehvar, M. T., and Navigli, R. (2016). Embeddings for word sense disambiguation: An evaluation study. In *Proceedings of the 54th Annual Meeting of the Association for Computational Linguistics (Volume 1: Long Papers)*, pages 897–907, Berlin, Germany, August. Association for Computational Linguistics.

Khapra, M. M., Shah, S., Kedia, P., and Bhattacharyya, P. (2010). Domain-specific word sense disambiguation combining corpus based and wordnet based parameters. In *In 5th International Conference on Global Wordnet (GWC2010)*.

Kolte, S. G. and Bhirud, S. G. (2008). Word sense disambiguation using wordnet domains. In *Emerging Trends in Engineering and Technology, 2008. ICETET'08. First International Conference on*, pages 1187–1191. IEEE.

Lesk, M. (1986). Automatic sense disambiguation using machine readable dictionaries: how to tell a pine cone from an ice cream cone. In *Proceedings of the 5th annual international conference on Systems documentation*, pages 24–26. ACM.

Magnini, B., Strapparava, C., Pezzulo, G., and Gliozzo, A. (2001). Using domain information for word sense disambiguation. In *The Proceedings of the Second International Workshop on Evaluating Word Sense Disambiguation Systems*, pages 111–114. Association for Computational Linguistics.

Martinez, D. (2005). Supervised word sense disambiguation: Facing current challenges. *Procesamiento del Lenguaje Natural*, 34.

Melamud, O., Goldberger, J., and Dagan, I. (2016). context2vec: Learning generic context embedding with bidirectional lstm. In *CoNLL*, pages 51–61.

Mikolov, T., Chen, K., Corrado, G., and Dean, J. (2013). Efficient estimation of word representations in vector space. *arXiv preprint arXiv:1301.3781*.

Miller, G. A. (1995). Wordnet: a lexical database for english. *Communications of the ACM*, 38(11):39–41.

Moro, A., Raganato, A., and Navigli, R. (2014). Entity linking meets word sense disambiguation: a unified approach. *Transactions of the Association for Computational Linguistics*, 2:231–244.

Navigli, R. (2009). Word sense disambiguation: A survey. *ACM Computing Surveys (CSUR)*, 41(2):10.

Navigli, R. (2012). A quick tour of word sense disambiguation, induction and related approaches. *SOFSEM 2012: Theory and practice of computer science*, pages 115–129.

Postma, M., Izquierdo, R., Agirre, E., Rigau, G., and Vossen, P. (2016). Addressing the mfs bias in wsd systems. In *LREC*.

Raganato, A., Camacho-Collados, J., and Navigli, R. (2017). Word sense disambiguation: A unified evaluation framework and empirical comparison. In *Proc. of EACL*, pages 99–110.

Rothe, S. and Schütze, H. (2015). Autoextend: Extending word embeddings to embeddings for synsets and lexemes. In *Proceedings of ACL*, pages 1793–1803.

Taghipour, K. and Ng, H. T. (2015a). One million sense-tagged instances for word sense disambiguation and induction. In *CoNLL*, pages 338–344.

Taghipour, K. and Ng, H. T. (2015b). Semi-supervised word sense disambiguation using word embeddings in general and specific domains. In *HLT-NAACL*, pages 314–323.

Taghipour, K. and Ng, H. T. (2015c). Semi-supervised word sense disambiguation using word embeddings in general and specific domains. In *HLT-NAACL*, pages 314–323.

Weissenborn, D., Hennig, L., Xu, F., and Uszkoreit, H. (2015). Multi-objective optimization for the joint disambiguation of nouns and named entities. In *ACL (1)*, pages 596–605.

Yuan, D., Richardson, J., Doherty, R., Evans, C., and Altendorf, E. (2016). Semi-supervised word sense disambiguation with neural models. In *COLING 2016*.

Zhong, Z. and Ng, H. T. (2010). It makes sense: A wide-coverage word sense disambiguation system for free text. In *Proceedings of the ACL 2010 System Demonstrations*, pages 78–83. Association for Computational Linguistics.

## 6. Language Resource References

Fondazione Bruno Kessler. (2007). *Labelling of WordNet 1.6 with semantic fields (WordNet Domains)*. 3.2, ISLRN 103-401-284-171-1.

Princeton University. (2006). *WordNet*. 3.0, ISLRN 379-473-059-273-1.